\documentclass[runningheads]{llncs}
\usepackage{graphicx}
\usepackage{booktabs, tabularx}
\usepackage{algpseudocode}
\usepackage{amssymb}
\usepackage{amsmath}
\usepackage{numprint}
\usepackage{tabularray}
\npthousandsep{\,}
\usepackage[hyphens]{url}
\usepackage{hyperref}
\usepackage[hyphenbreaks]{breakurl}
\usepackage{multirow}

\usepackage{bm}
\newcolumntype{C}{>{\centering\arraybackslash}X}

\begin{document}
\title{Learning Nuclei Representations with Masked Image Modelling}

\author{Piotr W\'ojcik\inst{1, 2} \and
Hussein Naji\inst{1, 2} \and
Adrian Simon\inst{4} \and
Reinhard Büttner\inst{4} \and
Katarzyna Bo\.zek\inst{1, 2, 3}}
\authorrunning{Piotr W\'ojcik et al.}
% First names are abbreviated in the running head.
% If there are more than two authors, 'et al.' is used.
%
\institute{Institute for Biomedical Informatics, Faculty of Medicine and University Hospital Cologne, University of Cologne, Germany \and
Center for Molecular Medicine Cologne (CMMC), Faculty of Medicine and University Hospital Cologne, University of Cologne, Germany \and
Cologne Excellence Cluster on Cellular Stress Responses in Aging-Associated Diseases (CECAD), University of Cologne, Germany \and
Institute of Pathology, University Hospital Cologne, Germany}
\maketitle
\begin{abstract}

Masked image modelling (MIM) is a powerful self-supervised representation learning paradigm, whose potential has not been widely demonstrated in medical image analysis. In this work, we show the capacity of MIM to capture rich semantic representations of Haemotoxylin \& Eosin (H\&E)-stained images at the nuclear level.
Inspired by Bidirectional Encoder representation from Image Transformers ({BEiT}) \cite{beit}, we split the images into smaller patches and generate corresponding discrete visual tokens. In addition to the regular grid-based patches, typically used in visual Transformers, we introduce patches of individual cell nuclei. We propose positional encoding of the irregular distribution of these structures within an image. We pre-train the model in a self-supervised manner on H\&E-stained whole-slide images of diffuse large B-cell lymphoma, where cell nuclei have been segmented. The pre-training objective is to recover the original discrete visual tokens of the masked image on the one hand, and to reconstruct the visual tokens of the masked object instances on the other.
Coupling these two pre-training tasks allows us to build powerful, context-aware representations of nuclei. Our model generalizes well and can be fine-tuned on downstream classification tasks, achieving improved cell classification accuracy on PanNuke dataset by more than $5\%$ compared to current instance segmentation methods. 

\keywords{nuclei classification  \and masked image modelling \and self-supervised learning.}
\end{abstract}

\section{Introduction}
In recent years, Transformer architectures \cite{transformer} have demonstrated the capability to achieve competitive results in computer vision \cite{vit}. However, their data-hungry training paradigm hinders many potential applications, especially in the field of digital pathology, where annotated data is limited. 
Several self-supervised solutions have been proposed to tackle this problem and proved to be very efficient in building good representations of visual data based on Transformers (e.g. \cite{dino}). The work of H. Bao \emph{et al.} \cite{beit} helped to bridge the gap between training paradigms used in natural language processing, e.g. in {BERT} \cite{bert}, and computer vision tasks. {BERT} randomly masks some word tokens and then reconstructs the missing tokens with the use of encoded representations of visible portions of the text. The fundamental difficulty in reproducing this pre-training schema for images is the magnitude
of potential vocabulary for even small image patches (e.g. $16 \times 16$ pixels). At the same time, large portion of the low-frequency visual detail is spurious for semantic decoding of an image.
%On the other hand, self-supervised methods that aim at the direct, pixel-wise reconstruction as their pretext task, suffer from prioritizing short-term visual dependencies over low-frequency ones which are semantically richer \cite{zeroshot}.
Bao et al. \cite{beit} addressed this problem by using a semantic-aware discrete image tokenizer \cite{zeroshot} which opened possibilities for efficient self-supervised learning of high-level, context-aware visual features via Transformers.
\\
\\
Concurrently, many approaches have been proposed to address a fundamental task in digital pathology -- segmentation and classification of nuclei. These approaches can roughly be categorised according to their backbone into Convolutional Neural Network (CNN)-based \cite{hovernet,rotation_cnn} and, more recently, Transformer-based \cite{transformer_segmentation}. For the task of nuclei segmentation, Generative Adversarial Network was proposed as a
method for synthesizing images with labels \cite{gan}. Despite promising advances in detection, classification of nuclei into various cell types is still considered as a by-product of instance segmentation and needs to be improved. Moreover, CNN-based methods are inherently limited by the locality of the view. This limitation poses a problem for pathological analysis, since the shape, size and density are not the sole indicators of a cell being cancerous. For example, in the diagnosis of diffuse large B-cell lymphoma (DLBCL), a highly heterogeneous disease both morphologically and clinically, neighboring cells as well as the subtle relationships between nucleus and cytoplasm are important to distinguish between healthy large leukocytes and lymphomatic cells. It is worth noting that several Transformer-based solutions have been proposed for learning good slide-level representations, with remarkable results achieved by R. Chen \emph{et al.} \cite{scale}, albeit no research known to us directly addresses the challenge of building context-aware representation of already segmented nuclei. 
\\
\\
Motivated by this need, we introduce a method for learning segmented nuclei representation which leverages the power of {BEiT}. Since its architecture is instance-agnostic, i.e. divides the image into a fixed number of regular grid-patches, we adopt the strategy presented in the work of S. Kim \emph{et al.} \cite{instaformer}. Namely, we incorporate instance-level patches of varying sizes, containing separate cell nuclei, into the sequence of patches fed to the backbone Transformer. We demonstrate that the representations of nuclei indeed reflect their respective cell types and require small amount of fine-tuning to achieve improved classification results compared to current state-of-the-art segmentation and classification methods.

\section{Method}
\begin{figure}[t]
\includegraphics[width=\textwidth]{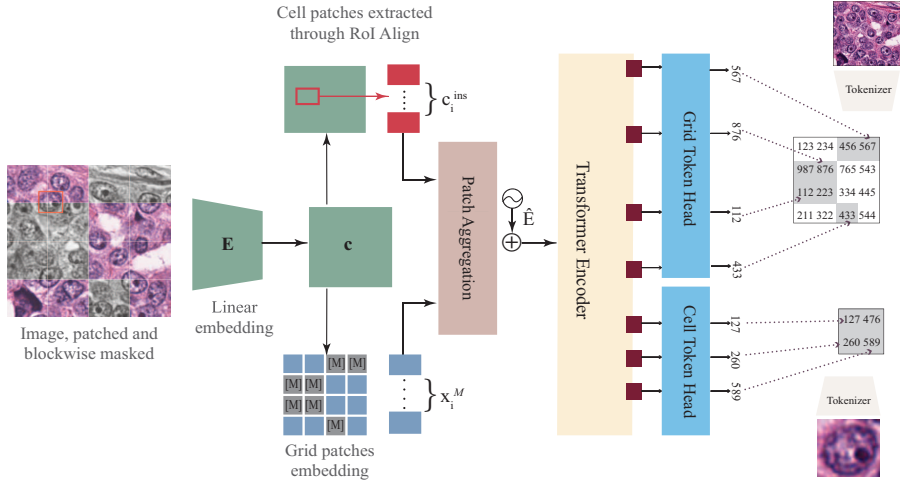}
\caption{Pipeline for self-supervised pre-training. We follow closely training schema proposed in \cite{beit}, which converts an image into discrete tokens. In addition to patches obtained by dividing an image into a regular grid we also tokenize parts of the image inside bounding boxes $B_i$, where each box encloses a single nucleus. We randomly mask a portion of an image, replacing masked patches (gray squares) with a special token \texttt{[M]}. We then embed image patches into a feature map $\mathbf{c}$. The same grid $\mathbf{c}$ is used to extract representations of nuclei through RoI Align \cite{roi_align}. Both grid patches and cell patches (represented by blue and red boxes, respectively) are fed into a vision Transformer backbone after aggregation. The pretext task aims at predicting discrete tokens for both input image and instance patches.} 
\label{fig:fig1}
\end{figure}

\subsection{Image Representation} 
As input, we are given an image $\bm{x}$ and bounding boxes of individual nuclei $B_i \in \mathcal{B}$, described by parameters $B_i = [x^{i}_{\min}, y^{i}_{\min}, x^{i}_{\max}, y^{i}_{\max}]$, i.e. by the coordinates of the upper-left and lower-right corners and $i \in 1, \ldots, N$ where $N$ is the total number of bounding boxes contained within an image $\bm{x}$. We apply the standard protocol for vision Transformers \cite{vit} which involves splitting a 2D image $\bm{x} \in \mathbb{R}^{H \times W \times C}$ into a sequence of grid patches $\bm{x}^p \in \mathbb{R}^{M \times (P^2 C)}$, where $C$ stands for the number of channels, $(H, W)$ represents the input size, $(P, P)$ is the size of a single patch and $L = HW/P^2$. In this work we run all experiments with $3$-channel images of size $448 \times 448$, divided into $28 \times 28$ square patches.

\paragraph{Grid Patch Embedding and Masking}
After obtaining grid patches, we use linear projection $\bm{E}\bm{x}^p$ (as depicted in Fig.~\ref{fig:fig1}) to embed grid patches into $D$-dimensional feature map $\bm{c}$, where $\bm{E} \in \mathbb{R}^{(P^2 C) \times D}$. Next, we randomly mask approximately $40\%$ of patches, employing blockwise masking described in detail in \cite{beit}. Indices of patches chosen for masking are denoted by $\mathcal{M}$. Then, we replace grid patch embeddings with a shared learnable token $\bm{e}_{[\mathcal{M}]}$ of a size $1 \times 1 \times D$: $\bm{e}^p_i$: $\bm{x}_i^{\mathcal{M}} = \delta(i \in \mathcal{M}) \odot \bm{e}_{[\mathcal{M}]} + (1 - \delta(i \in \mathcal{M})) \odot \bm{x}_{i}^p$, where $\delta(\cdot)$ is the indicator function. The choice of the masking algorithm and the ratio was motivated by ablation studies in \cite{beit}. Specifically, the use of the blockwise masking algorithm was shown to improve the model's accuracy compared to selecting patches for masking independently at random.

\paragraph{Cell Patch Embedding}
So far, we closely followed steps described in \cite{beit}. To encode not only the overall image structure but also the key semantic elements withing it - cells, we introduce the following modification to the baseline architecture inspired by the work \cite{instaformer}. Namely, given bounding boxes $B_i$ and the feature map $\bm{c}$ (see Fig.~\ref{fig:fig1}), we extract cell instance features through RoI Align module \cite{roi_align}:
\begin{equation}
\bm{c}^{\tiny{\textrm{ins}}}_{i} = \textrm{RoIAlign}(\bm{c}; B_i) \in \mathbb{R}^{k \times k \times D} 
\end{equation}
where $k \times k$ is the size of extracted features. In our work we set $k = 3$. Note that some cell instance features may be sampled from masked patches if their bounding boxes intersect with a masked portion of the image. Subsequently, we process the $\bm{c}^{\tiny{\textrm{ins}}}_{i}$ to obtain embedding vectors of dimension $1 \times 1 \times D$. We achieve this by applying  a convolution  on $\bm{c}^{\tiny{\textrm{ins}}}_{i}$, as proposed in \cite{instaformer}:
\begin{equation}
\bm{p}^{\tiny{\textrm{ins}}}_{i} = \textrm{Conv}(\bm{c}^{\tiny{\textrm{ins}}}_{i}) \in \mathbb{R}^{1 \times 1 \times D} 
\end{equation}

\paragraph{Patch Aggregation}
In our architecture, we concatenate a sequence of grid patch embeddings $\{\bm{c}^{\mathcal{M}}_{i}\}_{i=1}^{L}$ (blue squares in Fig.~\ref{fig:fig1}) and a sequence of $N$ cell patch embeddings $\{\bm{p}^{\tiny{\textrm{ins}}}_{i}\}_{i=1}^{N}$. Following \cite{beit}, we prepend a special, learnable token \texttt{[CLS]} to the concatenated sequence. Since the number of nuclei in a particular image may vary, we pad cell embedding sequences to a maximum length using \texttt{[PAD]} token. Overall, in this module we use three special tokens (\texttt{[M]}, \texttt{[PAD]}, \texttt{[CLS]}), each of them is shared, learnable and has size $1 \times 1 \times D$. After aggregation, the input sequence to the positional encoding module has the following form:
\begin{equation}
\bm{H}^0 = [\texttt{[CLS]}, \bm{c}^{\mathcal{M}}_{1}, \ldots, \bm{c}^{\mathcal{M}}_{L}, \underbrace{\bm{p}^{\tiny{\textrm{ins}}}_{1}, \ldots, \bm{p}^{\tiny{\textrm{ins}}}_{N}, \texttt{[PAD]}, \ldots, \texttt{[PAD]}}_\text{fixed maximum length}]
\end{equation}

\subsection{Positional Encoding}
\begin{figure}[t]
\includegraphics[width=\textwidth]{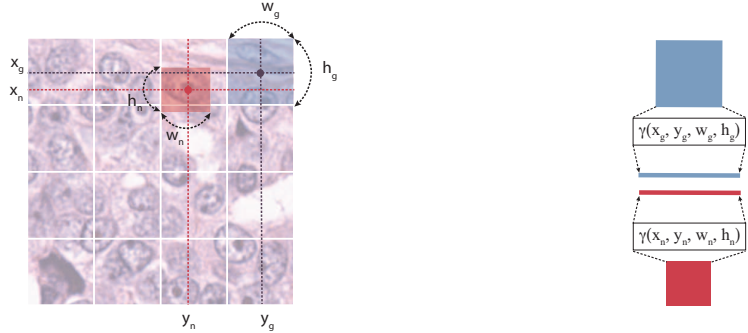}
\caption{Building positional encoding for both grid and cell patches. A tuple $(x_g, y_g, h_g, w_g)$ describes center coordinates, width and height of a grid patch (blue), while $(x_n, y_n, h_n, w_n)$ describes the position and shape of a cell patch (red).}
\label{fig:fig2}
\end{figure}

To encode the information about the positions of patches (both grid and cell), we use the technique proposed in \cite{instaformer}. The position and the shape of any patch, regardless of its regular grid or a bounding box identity, can be represented by center coordinates $(x, y)$ and patch width $w$ and height $h$, as exemplified in Fig.~\ref{fig:fig2}. 
We use a sinusoidal mapping $\gamma \colon \mathbb{R} \to \mathbb{R}^{D/4}$ defined as $\gamma(t) := [\sin(2^0\pi t), \cos(2^0\pi t), \ldots, \sin(2^{D/8-1}\pi t), \cos(2^{D/8-1}\pi t)]$.
For every patch token, positional encoding is constructed by concatenating all spatial information embedded by $\gamma$, namely $\bm{\hat{E}}_i = [\gamma(x_i), \gamma(y_i), \gamma(w_i), \gamma(h_i)]$. Positional encoding is further added to the corresponding token embedding: $\bm{H}_i^0 + \bm{\hat{E}}_i$. We handle $\texttt{[CLS]}$ token separately, with a learnable positional encoding of shape $1 \times 1 \times D$.

\subsection{Discrete Tokenizer}

The role of discrete tokenizer is to map an image to discrete tokens, \emph{i.e.} 
natural numbers. Formally, an image $\bm{x}$ is transformed into a grid $\bm{z} = [z_i, \ldots, z_T] \in \mathcal{V}^{(H/R) \times (W/R)}$, where $\mathcal{V}$ denotes the discrete vocabulary, $R$ is the resolution of the discrete tokenizer. In our work, we use pre-trained weights of a publicly available dVAE (\url{https://github.com/openai/DALL-E}) with $|\mathcal{V}| = 8192$. For the details of dVAE implementation and pre-training, please refer to \cite{dalle}. We tokenize the input image to into $28 \times 28$ grid. We then crop each nucleus instance to its bounding box and resize obtained crops to a size $32 \times 32$. Nuclei views are subsequently tokenized into $4 \times 4$ grids.  

\subsection{Backbone Transformer Encoder}

Our encoder shares all architectural details with ViT \cite{vit}, except for additional attention masking for preventing attending to \texttt{[PAD]} tokens, which we implement according to BERT \cite{bert}.
Embedded patches with positional encoding $(\bm{H}^0 + \bm{\hat{E}})$ are fed into $12$ layers of Transformer blocks with $12$ attention heads.
\begin{equation}
\bm{H}^{12} = [\bm{h}_{\texttt{[CLS]}}, \underbrace{\bm{h}^{g}_{1}, \ldots, \bm{h}^{g}_{28 \times 28}}_\text{\makebox[0pt]{\footnotesize grid representations}}, \overbrace{\bm{h}^{n}_{1}, \ldots, \bm{h}^{n}_{N}}^\text{\makebox[0pt]{\footnotesize nuclei representations}},  \bm{h}_{\texttt{[PAD]}}, \ldots, \bm{h}_{\texttt{[PAD]}}].
\end{equation}
After discarding the representation of $\texttt{[CLS]}$ and $\texttt{[PAD]}$ tokens, $\bm{h}_{\texttt{[CLS]}}$ and $\bm{h}_{\texttt{[PAD]}}$ respectively, we are left with $28 \times 28$ 
representations of regular grid patches $\bm{h}^{g}_{i}$ and $N$ representations of nuclei $\bm{h}^{n}_{j}$.
\subsubsection{Pre-training Loss Function}
We use two fully-connected layers as output heads (see Fig.~\ref{fig:fig1}). The first one, image head, predicts discrete tokens from the representation of every masked grid patch $\{\bm{h}^{g}_{i}: i \in \mathcal{M}\}$. Since the input image was tokenized into $28 \times 28$ natural numbers, there is one-to-one correspondence between grid patches and tokens. The second,  cell instance head, predicts $4$ tokens from $\bm{h}^{n}_{j}$, for each cell instance whose bounding box $B_j$ intersects with masked portion of the image. The training MIM loss is defined as:
\begin{equation}
\mathcal{L}_{\text{MIM}} = -\sum_{(\bm{x}, \mathcal{B}) \in \mathcal{D}} \left(\overbrace{ \sum_{i \in \mathcal{M}} \log(p(z^g_i|\bm{c_i}^{\mathcal{M}}))}^\text{\makebox[0pt]{\footnotesize BEiT loss}} + \overbrace{\sum_{j \in \mathcal{B}_{\mathcal{M}}} \sum_{k=1}^{4} \log(p(z^n_{j,k}|\bm{p_j}^{\text{ins}})}^\text{\makebox[0pt]{\footnotesize $\mathcal{L}_{\text{inst}}$}}) \right)   
\end{equation}
where $\mathcal{D}$ denotes pre-training set of images with bounding boxes, $\mathcal{B}_{\mathcal{M}}$ indicates the set $\mathcal{B}$ of bounding boxes that intersect with masked grid patches, $z^g_i$ the discrete token for the $i$-th grid patch and $z^n_{j,1}, \ldots, z^n_{j, 4}$ discrete tokens for $j$-th nuclei instance. Notice, that the first component of the loss function comes directly from \cite{beit} while the second, dubbed $\mathcal{L}_{\text{inst}}$, is proposed by us in order to sharpen the view on nuclei.
\section{Experiments}
\begin{figure}[t]
\includegraphics[width=\textwidth]{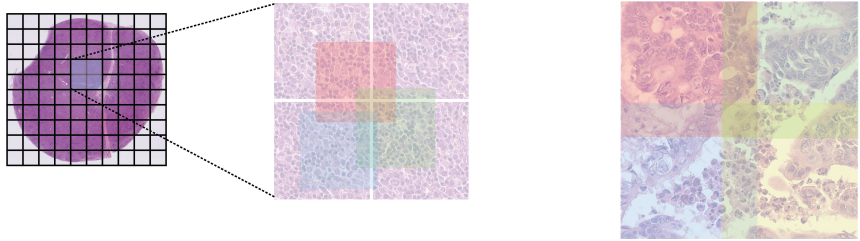}
\caption{Datasets augmentation strategy. \textbf{Left}: Self-supervised pre-training dataset consists of one H\&E-stained WSI of a lymph node, divided into tiles of size $512 \times 512$. For each four adjacent tiles, we randomly generate overlapping crops. \textbf{Right}: To evaluate our model on the CoNSeP dataset, we divide $1000 \times 1000$ pixel-large images into four overlapping tiles of $448 \times 448$. Each nucleus is classified basing on its representation from one tile only.}
\label{fig:fig3}
\end{figure}
\paragraph{Datasets}
We used three datasets in our work. For self-supervised pre-training, we used our in-house dataset of $\numprint{37665}$ H\&E images ($40$x magnification) obtained from a single WSI of a DLBCL lymph node. Each $512 \times 512$ tile was segmented with a bounding box for each nucleus, albeit no cell labels were provided. The segmentation was performed automatically, without further verification. We resized every tile to $448 \times 448$ and randomly cropped additional tiles to increase the number of training examples, as demonstrated in Fig.~\ref{fig:fig3}. After preprocessing, DLBCL dataset consists of $\numprint{160545}$ image tiles with segmented nuclei. For fine-tuning and testing we use CoNSeP and PanNuke datasets \cite{hovernet,pannuke}. The CoNSeP dataset consists of $\numprint{24319}$ annotated nuclei from $41$ H\&E images. Nuclei are grouped into four types. The PanNuke dataset \cite{pannuke} contains $\numprint{205343}$ annotated nuclei of five types. Images in both datasets have size of $256 \times 256$ and originate from $19$ different tissue types. We did not perform any pre-processing on these images except for resizing them to $448 \times 448$.
\paragraph{Pre-Training Setup}
The proposed network was pre-trained with similar set of parameters as
BEiT \cite{beit}. We used $12$-layer Transformer with $768$ hidden size and $12$ attention heads. We pre-trained the model for $800$ epochs with a batch size $96$. The codebase used for experiments was that of the original BEiT implementation \url{https://github.com/microsoft/unilm/tree/master/beit}. Instance embedding code was built upon InstaFormer implementation \url{https://github.com/KU-CVLAB/InstaFormer}. Around $550$-th epoch, we added PanNuke Fold $1$ dataset and CoNSeP training dataset to the initial DLBCL dataset to boost performance. Throughout the pre-training phase, we applied the same set of augmentations as in \cite{beit}, which included RandomResizeAndCrop, color jittering with a parameter of 0.4 and RandomFlip. Additionally, all input images were normalized to the mean and standard deviation of ImageNet.
\paragraph{Fine-tuning on Annotated Datasets}
We fine-tuned our model and validated its performance on the two labelled datasets, CoNSeP and PanNuke. In nuclei classification task, we discarded two MIM pre-training linear layers and used softmax classifier on nuclei representations: $\text{softmax}(\{\bm{h}_{i}^{n}\}^{N}_{i=1} \bm{W}_C)$, where $\bm{h}_{i}^{n}$ stands for $i$-th nuclei representation, $C$ is the number of nuclei classes and $\bm{W}_C$ denotes a linear layer parameter matrix.
\paragraph{Evaluation Protocol}
As a baseline model, we chose the widely used state-of-art nuclei detection and classification model HoVer-Net \cite{hovernet}. Since the authors only provided metrics relevant to the combined tasks of nuclei detection and classification, for a fair comparison, we modified the codebase to print both accuracy and F1 score for the subset of nuclei that were correctly segmented. Results presented in Tab.~\ref{tab1} were obtained with a use of published pre-trained weights \url{https://github.com/vqdang/hover_net} for the CoNSeP and PanNuke datasets. Subsequently, we ran evaluation of our model on the exact shapes of correctly detected nuclei.
\begin{figure}
\includegraphics[width=\textwidth]{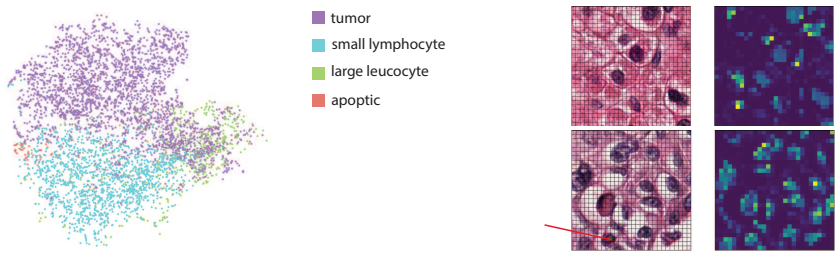}
\caption{\textbf{Left}: t-SNE visualisation of DLBCL nuclei representations without fine-tuning on any labeled dataset.  \textbf{Right}: Upper row shows self-attention map for \texttt{[CLS]} token; lower row shows self-attention map for a reference point indicated with an arrow.}
\label{fig:fig4}
\end{figure}
\paragraph{Ablation Studies on $\mathcal{L}_{\text{inst}}$} We performed ablation studies to test the importance
of $\mathcal{L}_{\text{inst}}$. We pre-trained two variants of the model (with and w/o $\mathcal{L}_{\text{inst}}$) for $250$ epochs on the DLBCL dataset and compared linear probing results on Fold 3 of PanNuke dataset. We kept the models frozen and added a linear head to evaluate the self-supervised models \cite{beit}. The model pre-trained using the full $\mathcal{L}_{\text{MIM}}$ loss function achieved a significantly higher accuracy of $0.613$ compared to the model pre-trained with only the BEiT component of the loss function, which achieved an accuracy of $0.517$.
\section{Results and Conclusion}
As demonstrated in Tab.~\ref{tab1}, our model visibly outperforms HoVerNet in the task of nuclei classification. The difference is especially pronounced on the PanNuke dataset ($0.05$ on average) for all dataset which is much larger than CoNSeP. We find striking the capacity of our model to generalize to images from a pathological domain completely different from the DLBCL corpus used for pre-training. 
Fig.~\ref{fig:fig4} shows the ability of self-supervised training to separate cells of different types while self-attention map demonstrates that our model learns long-distance relations between nuclei. Although our proposed training method requires a non-negligible amount of segmented images, these can be obtained automatically at a low cost. It should be also noted that MIM methods achieve competitive results on image segmentation \cite{beit}, offering new possibilities for performing both tasks of nuclei detection and segmentation simultaneously in the future. 
\begin{table}
\caption{Nuclei classification results on PanNuke dataset (Fold 1 - Fold 3 split) and CoNSeP. HoVerNet (HN) is used as a baseline model.}\label{tab1}
\begin{tabular}{ p{2.0cm}p{1.60cm} p{1.60cm}p{1.60cm} p{1.60cm} p{1.60cm} p{1.60cm}  }
 \hline
 \multicolumn{7}{c}{\textbf{PanNuke}} \\
\hline
 & \multicolumn{1}{l}{\textbf{All}} & \multicolumn{1}{l}{\textbf{Neo.}} &  \multicolumn{1}{l}{\textbf{Inflam.}} & \multicolumn{1}{l}{\textbf{Conn.}} & \multicolumn{1}{l}{\textbf{Dead}} & \multicolumn{1}{l}{\textbf{Epith.}} \\
 \cline{2-7}
HN F1  & $0.780$ & $0.856$ &  $0.714$ & $0.719$ & $0.399$  & $0.765$\\
 Ours F1 &   $\mathbf{0.831}$  & $\mathbf{0.893}$  &$\mathbf{0.776}$ & $\mathbf{0.764}$ & $\mathbf{0.571}$ & $\mathbf{0.848}$\\
\hline
 HN Acc  & $0.780$   & $0.748$&  $0.555$ &  $0.562$ & $0.249$ & $0.619$ \\
 Ours Acc &   $\mathbf{0.832}$  & $\mathbf{0.807}$  & $\mathbf{0.634}$ & $\mathbf{0.618}$ & $\mathbf{0.400}$ & $\mathbf{0.735}$\\
  \hline
 \multicolumn{7}{c}{\textbf{CoNSeP}} \\
\hline
 & \multicolumn{1}{l}{\textbf{All}} & \multicolumn{1}{l}{\textbf{Misc.}} & \multicolumn{1}{l}{\textbf{Inflam.}} & \multicolumn{1}{l}{\textbf{Epith.}} & \multicolumn{1}{l}{\textbf{Spindle}} \\
 \cline{2-7}
  HN F1   & $0.872$    & $0.734$ &  $0.817$ &  $0.939$  & $0.856$\\
 Ours F1 &   $\mathbf{0.885}$  & $\mathbf{0.804}$  &$\mathbf{0.822}$ & $\mathbf{0.954}$ & $\mathbf{0.864}$\\
 \hline
 HN Acc  & $0.872$    & $0.580$&  $0.690$ &  $0.884$ & $0.749$ \\
 Ours Acc &   $\mathbf{0.885}$  & $\mathbf{0.672}$  & $\mathbf{0.698}$ & $\mathbf{0.911}$ & $\mathbf{0.761}$\\
 \hline
 \end{tabular}
\end{table}

\bibliographystyle{splncs04}
\bibliography{nuclei_classification_mim}

\end{document}